%% file: main.tex
\documentclass[10pt,twocolumn,letterpaper]{article}

\usepackage{wacv}
\usepackage{times}
\usepackage{epsfig}
\usepackage{graphicx}
\usepackage{amsmath}
\usepackage{amssymb}
\usepackage{booktabs}
\def\wacvPaperID{850} % Enter the WACV Paper ID here

\wacvalgorithmstrack   % Uncomment this line if you are submitting to the Algorithms Track.
\wacvfinalcopy % *** Uncomment this line for the final submission
\ifwacvfinal
\fi
\ifwacvfinal
\usepackage[breaklinks=true,bookmarks=false]{hyperref}
\else
\usepackage[pagebackref=true,breaklinks=true,colorlinks,bookmarks=false]{hyperref}
\fi
\input{content/preamble.tex}

\ifwacvfinal
\else
\fi

\newcommand{\customfootnotetext}[2]{{% Group to localize change to footnote
  \renewcommand{\thefootnote}{#1}% Update footnote counter representation
  \footnotetext[0]{#2}}}% Print footnote text

\begin{document}

%%%%%%%%% TITLE
\title{InterTrack: Interaction Transformer for 3D Multi-Object Tracking}

\author{
John Willes\textsuperscript{*} \ \ \ \ Cody Reading\textsuperscript{*} \ \ \ \ Steven L. Waslander\\
University of Toronto Robotics Institute\\
{\tt\small \{john.willes, cody.reading\}@mail.utoronto.ca, steven.waslander@robotics.utias.utoronto.ca }
}

\maketitle
\customfootnotetext{*}{indicates equal contribution}
%\thispagestyle{empty}

%%%%%%%%% ABSTRACT
\input{content/abstract}

%%%%%%%%% BODY TEXT
\input{content/intro}
\input{content/related_work}
\input{content/methodology}
\input{content/results}
\input{content/conclusion}

%%%%%%%%% REFERENCES

{\small
\bibliographystyle{ieee_fullname}
\bibliography{intertrack}
}
 
 \clearpage
 
\appendix 
\input{content/appendix}

\end{document}

%% file: content/preamble.tex
\makeatletter
\@namedef{ver@everyshi.sty}{}
\newcommand\newtag[2]{#1\def\@currentlabel{#1}\label{#2}}
\makeatother
\usepackage{svg}
\usepackage{xcolor,colortbl}
\usepackage{booktabs}
\usepackage[super]{nth}
\usepackage{siunitx}
\usepackage{multirow}
\usepackage{amsmath}
\usepackage{hyperref} 
\usepackage{fancyhdr}

\newcommand{\method}{InterTrack}
\definecolor{ImproveBlue}{HTML}{E0FFFF}
\definecolor{FirstRed}{HTML}{FE0000}
\definecolor{SecondBlue}{HTML}{3531FF}
\newcolumntype{x}[1]{>{\centering\arraybackslash\hspace{0pt}}p{#1}}

\DeclareMathOperator*{\argmin}{arg\,min}

%% file: content/abstract.tex
\begin{abstract}
    3D multi-object tracking (MOT) is a key problem for autonomous vehicles, required to perform well-informed motion planning in dynamic environments. Particularly for densely occupied scenes, associating existing tracks to new detections remains challenging as existing systems tend to omit critical contextual information. Our proposed solution, \method, introduces the Interaction Transformer for 3D MOT to generate discriminative object representations for data association. We extract state and shape features for each track and detection, and efficiently aggregate global information via attention. We then perform a learned regression on each track/detection feature pair to estimate affinities, and use a robust two-stage data association and track management approach to produce the final tracks. We validate our approach on the nuScenes 3D MOT benchmark, where we observe significant improvements, particularly on classes with small physical sizes and clustered objects. As of submission, \method~ranks $1^{st}$ in overall AMOTA among methods using CenterPoint~\cite{CenterPoint} detections.
\end{abstract}

%% file: content/intro.tex
\section{Introduction} \label{sec:intro}
%-----------------------------------------------------------------------
\begin{figure}[t]
\begin{center}
\includegraphics[width=\linewidth]{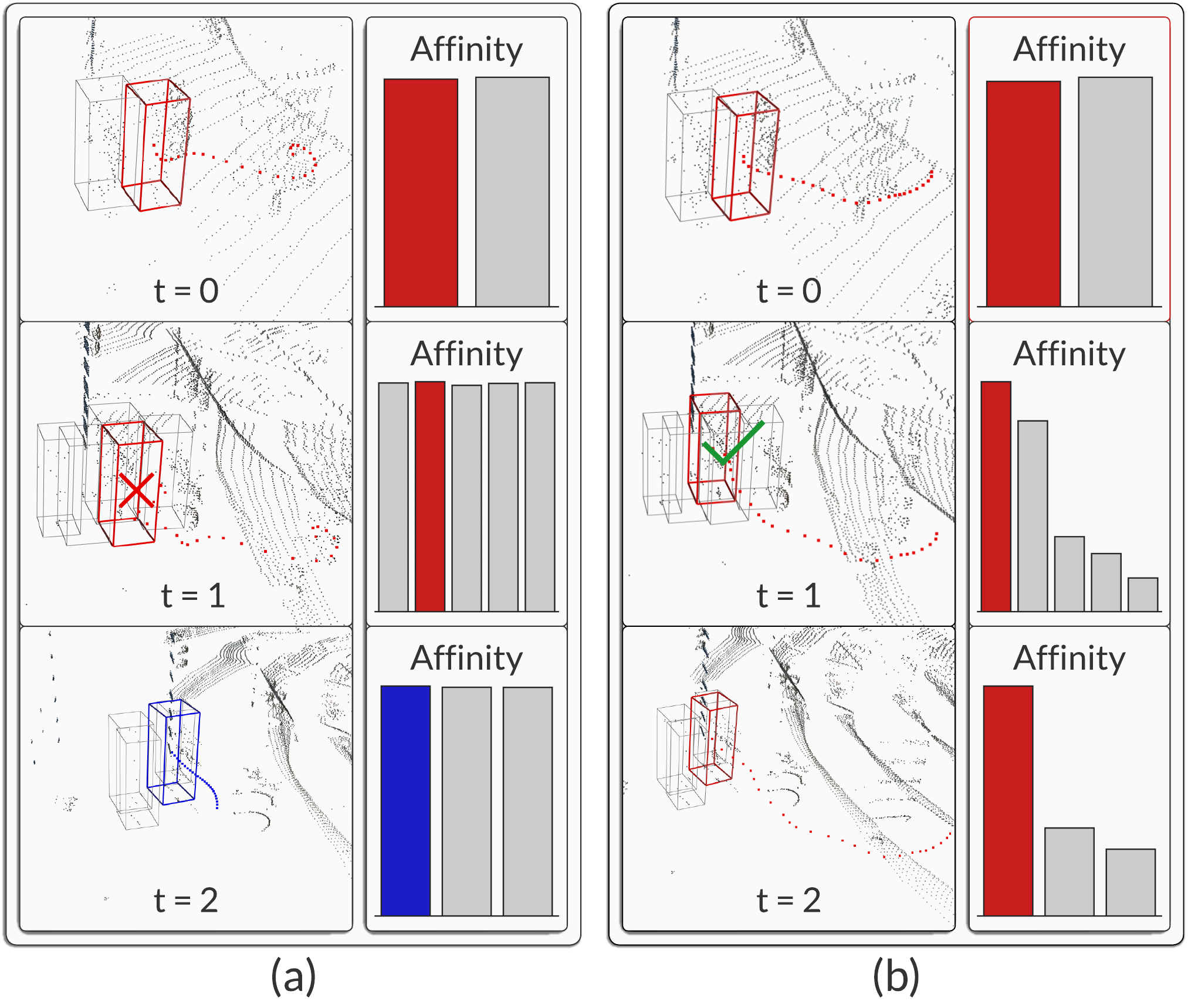}
\end{center}
 \caption{3D tracking visualization. (a) When object interactions are not considered, the highlighted track produces multiple similar affinity scores leading to an incorrect match at $t = 1$. The track is killed at $t = 2$, resulting in an identity switch (shown as new color). (b) Interaction modeling encourages a single highest affinity for the correct detection resulting in an uninterrupted track.}
\label{fig:intro}
\end{figure}
%-------------------------------------------------------------------------
3D multi-object tracking (MOT) is a vital task for many fields such as autonomous vehicles and robotics, enabling systems to understand their environment in order to react accordingly. A common approach is the tracking-by-detection paradigm~\cite{AB3DMOT,CenterPoint,EagerMOT,ProbMOTv1}, in which trackers consume independently generated 3D detections as input. Existing tracks are matched with new detections at each frame, by estimating a track/detection affinity matrix and matching high affinity pairs. Incorrect associations can lead to identity switching and false track initialization, confusing decision making in subsequent frames. 

Affinities are estimated by extracting track and detection features, followed by a pairwise comparison to estimate affinity scores. Feature extraction can simply extract the object state~\cite{AB3DMOT,EagerMOT} or use a neural network to extract features from sensor data~\cite{PTP,ProbMOTv2}. For effective association, features should be discriminative such that feature comparison results in accurate affinity estimation~\cite{GNN3DMOT}.

3D MOT methods often perform feature extraction independently for each track and detection. Independent methods, however, tend to suffer from high feature similarity, particularly for densely clustered objects (see Figure~\ref{fig:intro}). Similar features are non-discriminative, and lead to ambiguous data association as the correct match cannot be distinguished from the incorrect match hypotheses. 

Feature discrimination can be improved through learned feature interaction modeling~\cite{GNN3DMOT}. Interaction modeling allows individual features to affect one another, adding a capacity for the network to encourage discrimination between matching and non-matching feature pairs. Previous methods~\cite{GNN3DMOT,PTP} have adopted a graph neural network (GNN) to model feature interactions, but limit the connections to maintain computational efficiency. We argue all interactions are important, as for example, short range interactions can be helpful for differentiating objects in dense clusters, while long range interactions can assist with fast-moving objects with large motion changes.

Additionally, most prior 3D MOT works~\cite{CenterPoint,ProbMOTv2} lack a filtering of duplicate tracks, leading to additional false positives and reduced performance. Duplicate tracks can be an issue for trackers that ingest detections~\cite{CenterPoint} with high recall and many duplicate false positives.

To resolve the identified issues, we propose \method, a 3D MOT method that introduces the Interaction Transformer to 3D MOT, in order to generate discriminative affinity estimates in an end-to-end manner, and introduces a track rejection module. We summarize our approach with three contributions.
%------------------------------------------------------------------------
\\[0.2\baselineskip]
\noindent\textbf{(1) Interaction Transformer}. We introduce the Interaction Transformer for 3D MOT, a learned feature interaction mechanism leveraging the Transformer model. The Interaction Transformer models spatio-temporal interactions between all object pairs, leveraging attention layers~\cite{Attention} to maintain high computational efficiency. Through complete interaction modeling, we encourage feature discrimination between all object combinations, leading to increased overall feature discrimination and tracking performance shown in Section~\ref{sec:ablations}. 
%------------------------------------------------------------------------
\\[0.2\baselineskip]
\noindent\textbf{(2) Affinity Estimation Pipeline}. We design a novel method to estimate track/detection affinities in an end-to-end manner. Using detections and LiDAR point clouds, our method extracts full state and shape information for each track and detection, and aggregates contextual information via the Interaction Transformer. Each track/detection feature pair is used to regress affinity scores. Our learned feature extraction offers increased representational power over simply extracting the object state and allows for interaction modeling. 
%------------------------------------------------------------------------
\\[0.2\baselineskip]
\noindent\textbf{(3) Track Rejection}. We introduce a duplicated track rejection strategy, by removing any tracks that overlap greater than a specified 3D intersection-over-union (IoU) threshold. We validate the track rejection module leads to performance improvements as demonstrated in Section~\ref{sec:ablations}.
%------------------------------------------------------------------------
\\[0.2\baselineskip]
\indent
\method~is shown to rank \nth{1} on the nuScenes 3D MOT test benchmark~\cite{nuscenes2019} among methods using public detections, with margins of 1.00\%, 2.20\%, 4.70\%, and 2.60\% AMOTA on the Overall, Bicycle, Motorcycle, and Pedestrian categories, respectively.
%------------------------------------------------------------------------

%% file: content/related_work.tex
\section{Related Work} \label{sec:RelatedWork}
%------------------------------------------------------------------------
\begin{figure*}[t]
\begin{center}
\includegraphics[width=\textwidth]{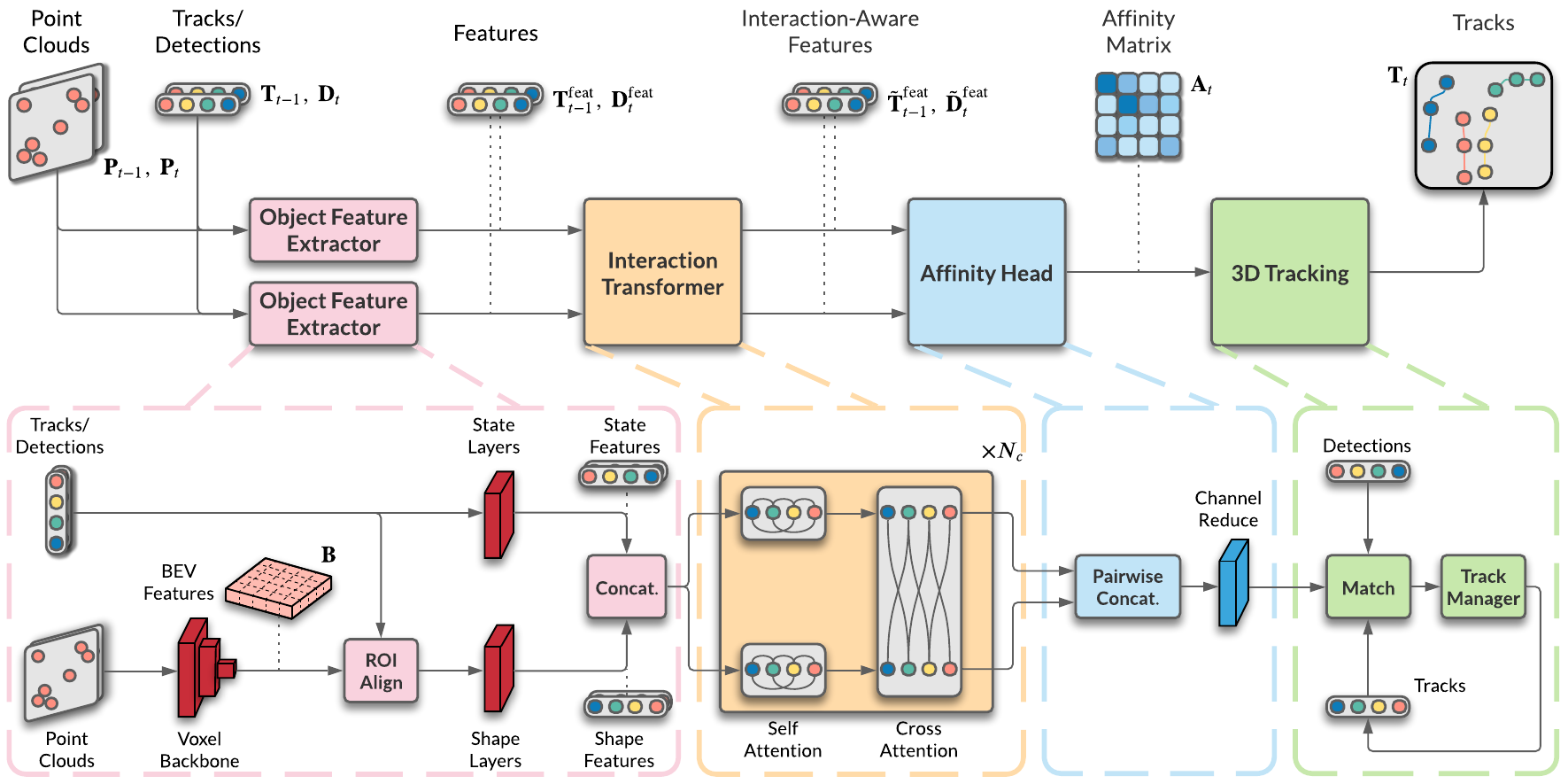}
\end{center}
   \caption{\method~Architecture. For each frame, track and detection features $\mathbf{T}_{t-1}^{\mathrm{feat}},~\mathbf{D}_t^{\mathrm{feat}}$ are extracted from object and frame level information, followed by a Transformer module to produce interaction-aware track/detection features $\tilde{\mathbf{T}}_{t-1}^{\mathrm{feat}},~\tilde{\mathbf{D}}_t^{\mathrm{feat}}$. The track/detection features are then used to estimate an affinity matrix $\mathbf{A}_t$, which is consumed by a 3D tracking module to produce the final tracks $\mathbf{T}_t$.}
\label{fig:architecture}
\end{figure*}
%------------------------------------------------------------------------
\noindent
\textbf{Transformer}. 
The Transformer~\cite{Attention} architecture has become the standard sequence modeling technique for natural language processing (NLP) tasks~\cite{BERT,RoBERTa,GPT-3}. The key element is the attention mechanism used to model dependencies between all sequence elements, leading to improved computational efficiency and performance. Transformers have seen use in computer vision tasks~\cite{imagewords,DETR,LoFTR}, in which image features are extracted to be consumed by a Transformer as input. Transformers have also seen use in 3D prediction~\cite{InterPred} wherein the Transformer operates on object-level features extracted from autonomous vehicle sensor input. In this work, we use the Transformer to operate on object-level features, and leverage the attention mechanism to model interactions between all tracks and detections.
%------------------------------------------------------------------------
\\[0.2\baselineskip]
\noindent
\textbf{Image Feature Matching}. 
Image feature matching consists of matching points between image pairs, often used in applications such as simultaneous localization and mapping (SLAM) and visual odometry. Feature descriptors are extracted from local regions surrounding each point, in order to match points with the highest feature similarity. Feature extraction can be performed by hand-crafting features~\cite{SIFT,BRIEF,ORB} or with a learned convolutional neural network (CNN)~\cite{UCN,DELF}. SuperGlue~\cite{SuperGlue} and LoFTR~\cite{LoFTR} introduce a GNN and Transformer module respectively, as a means to incorporate both local and global image information during feature extraction. We follow the Transformer design of LoFTR~\cite{LoFTR} to incorporate global object information in our feature extraction.
%------------------------------------------------------------------------
\\[0.2\baselineskip]
\noindent
\textbf{Multi-Object Tracking}. MOT is a key focus in both 2D~\cite{2DMOTSurvey} and 3D~\cite{AB3DMOT} settings. Tracking methods either follow a tracking-by-detection approach~\cite{AB3DMOT,SORT,FAMNet}, wherein object detections are independently estimated and consumed by a tracking module for track estimation, or perform joint detection and tracking in a single stage~\cite{SimTrack,TrackFormer,RetinaTrack,TransTrack}. 

In 2D MOT, recent methods~\cite{TransTrack,TrackFormer,MOTR} leverage the Transformer for joint detection and tracking, extending DETR~\cite{DETR} to associate detections implicitly. A drawback is the lack of explicit affinity estimates, resulting in a less interpretable solution often required by safety-critical systems.

Most 3D MOT methods~\cite{AB3DMOT,AlphaTrack,ProbMOTv1,ProbMOTv2} implement the tracking-by-detection paradigm, which is dependent on high-quality, high-frequency detections~\cite{CenterPoint,CBGS}. These trackers solve a data association problem~\cite{GlobalDA,Mono-Camera3D}, in which detections are assigned to existing tracks or used to birth new tracks. Similar to feature matching methods, tracking-by-detection methods extract object features and form correspondences with the highest feature similarity~\cite{GlobalGreedy,JointMono3D}. Affinity matrices are estimated that represent the feature similarity between tracks and detections, and are fed to either a Greedy or Hungarian~\cite{Hungarian} matching algorithm to generate matches. Based on how data association is performed, trackers can be divided into two categories: heuristic and learned association methods.
%------------------------------------------------------------------------
Heuristic-based methods compute affinity scores using a heuristic similarity or distance metric. The metric is based on the object state, with some metrics including 3D IoU, Euclidean distance, or Mahalanobis distance~\cite{mahalanobis}. One of the first methods for 3D tracking is AB3DMOT~\cite{AB3DMOT}, which uses 3D IoU as a similarity metric and a Kalman filter for track prediction and update. Introduced by CenterPoint~\cite{CenterPoint}, methods~\cite{TransFusion, BEVFusion} can compute affinity using Euclidean distance, and replaces the Kalman filter prediction with motion prediction based on learned velocity estimates. EagerMOT~\cite{EagerMOT} performs separate data association stages for LiDAR and camera streams, and modifies the Euclidean distance to include orientation difference. Without contextual information, heuristic-based affinity estimation leads to many similar affinity scores and ambiguous data association, notably for clustered objects.

Learning-based methods use a neural network to extract object features from detection and sensor data. Object affinities are computed from the extracted object features using a distance metric~\cite{AlphaTrack} or a learned scoring function~\cite{GNN3DMOT,ProbMOTv2}. GNN3DMOT~\cite{GNN3DMOT} and PTP~\cite{PTP} aggregate object state and appearance information from LiDAR and image streams to be consumed by a GNN to model object interactions. The object features are represented as GNN nodes that are passed to an edge regression module to estimate affinities. Edges are sparsely constructed due to computational requirements of the GNN, where edges are only formed between tracks and detections if the objects are within specified 2D and 3D distance thresholds. Chiu et al.~\cite{ProbMOTv2} similarly extract object features from LiDAR and image data, but simply apply a pairwise concatenation and regression to estimate affinities, which are combined with the Mahalanobis distance~\cite{mahalanobis} to generate affinity scores. AlphaTrack~\cite{AlphaTrack} improves the CenterPoint~\cite{CenterPoint} detector by fusing LiDAR point clouds with image features, and extracts object appearance information for data association. SimTrack~\cite{SimTrack} performs data association implicitly in a learned joint detection and tracking framework, based on reading off identities from the prior frame's centerness map. Current learned association methods either extract features independently or with sparse interaction modeling, limiting feature discrimination and region of influence that is considered during matching.  

\method~directly address the limitation of interaction modeling for data association, by introducing the Interaction Transformer to 3D MOT. Doing so allows for complete interaction modeling, leading to improved object feature discrimination. \method~only leverages the Transformer for data association to provide an interpretable solution for safety-critical applications in autonomous vehicles.
%------------------------------------------------------------------------

%% file: content/methodology.tex
\section{Methodology} \label{sec:Method}
%------------------------------------------------------------------------
\method~learns to estimate affinity matrices for data association, followed by a 3D tracking module to estimate tracks. An overview of \method~is shown in Figure~\ref{fig:architecture}.
%------------------------------------------------------------------------
\subsection{Affinity Learning} \label{sec:affinity-learning}
%------------------------------------------------------------------------
Our network learns to estimate discriminative affinity matrices for effective data association. Taking object detections and LiDAR point clouds as input, we extract object-level features for both existing tracks and new detections. Interaction information is embedded in the track/detection features with a Transformer model, followed by an affinity estimation module to regress affinity scores.
%------------------------------------------------------------------------
\\[0.5\baselineskip]
\noindent\textbf{Object Feature Extraction}. We extract features for both tracks and detections, using a shared network to extract both state and shape features for each object. The inputs to the feature extractor are either track or detection states, $\mathbf{T}_{t-1} \in \mathbb{R}^{M \times 11}$ or $\mathbf{D}_{t} \in \mathbb{R}^{N \times 11}$, and frame corresponding LiDAR point clouds, $\mathbf{P}_{t-1} \in \mathbb{R}^{P \times 5}$ or $\mathbf{P}_{t} \in \mathbb{R}^{P \times 5}$. We extract state directly as $(x, y, z, l, w, h, \theta, \dot{x}, \dot{y}, c, s)$ including 3D location $(x, y, z)$, dimensions $(l, w, h)$, heading angle $\theta$, velocity $(\dot{x}, \dot{y})$, class $c$, and confidence $s$, while shape information is extracted from the LiDAR point cloud, $\mathbf{P}$. We adopt the point cloud preprocessing and 3D feature extractor used in SECOND~\cite{SECOND} (Voxel Backbone in Figure~\ref{fig:architecture}) to generate bird's-eye-view (BEV) features, $\mathbf{B}$, where multiple LiDAR sweeps are combined and each point is represented as $(x, y, z, r, \Delta t)$ including 3D location $(x, y, z)$, reflectance $r$, and relative timestep $\Delta t$. 3D object states are projected into the BEV grid, $\mathbf{B}$, and used with ROI Align to extract shape features for each object. Let $\mathbf{G}(\cdot)$ represent a feed-forward network (FFN) consisting of 2 Linear-BatchNorm-Relu blocks, used to modify the number of feature channels. We apply a state FFN, $\mathbf{G}_{\mathrm{st}}$, and a shape FFN, $\mathbf{G}_{\mathrm{sh}}$ (State and Shape Layers in Figure~\ref{fig:architecture}) to modify the number of channels to $C/2$, and then concatenate to form track or detection features, $\mathbf{T}_{t-1}^{\mathrm{feat}} \in \mathbb{R}^{M \times C}$ or $\mathbf{D}_{t}^{\mathrm{feat}} \in \mathbb{R}^{N \times C}$.
%------------------------------------------------------------------------
\begin{figure}[t]
\begin{center}
\includegraphics[width=\linewidth]{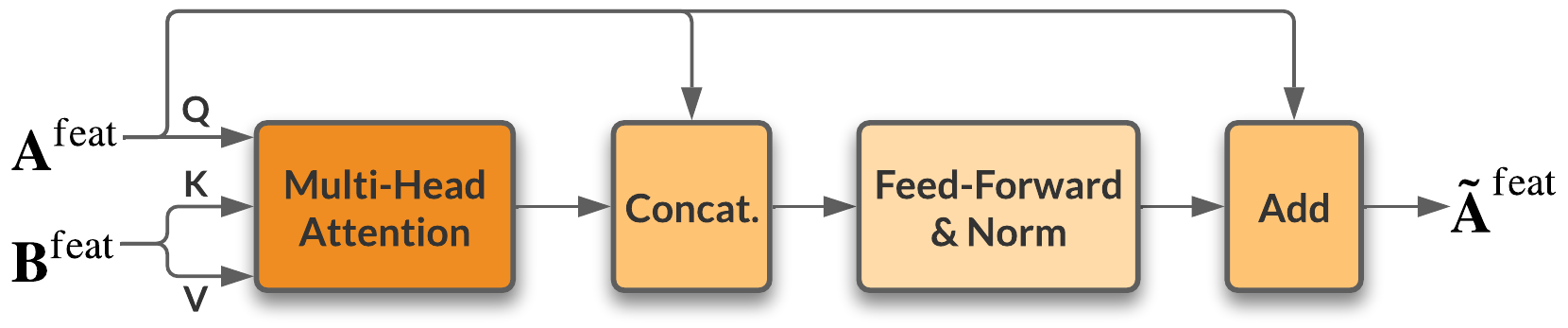}
\end{center}
   \caption{Attention block. $\mathbf{A}^{\mathrm{feat}}, \mathbf{B}^{\mathrm{feat}}$ is the input feature pair and attention is based on the original Multi-Head Attention~\cite{Attention}.}
\label{fig:attention}
\end{figure}
%------------------------------------------------------------------------
\\[0.2\baselineskip]
\noindent\textbf{Interaction Transformer}.
The track and detection features, $\mathbf{T}_{t-1}^{\mathrm{feat}}$ and $\mathbf{D}_t^{\mathrm{feat}}$, are used as input to the Interaction Transformer (see Figure~\ref{fig:architecture}), producing interaction-aware track and detection features, $\mathbf{\tilde{T}}_{t-1}^{\mathrm{feat}} \in \mathbb{R}^{M\times C}$ and $\mathbf{\tilde{D}}_t^{\mathrm{feat}} \in \mathbb{R}^{N\times C}$. We adopt the transformer module from LoFTR~\cite{LoFTR} and interleave the self and cross attention blocks $N_c$ times. In this work, we retain the value of $N_c = 4$.

The design of the attention block is shown in Figure~\ref{fig:attention}. For self-attention blocks, the input features, $\mathbf{A}^{\mathrm{feat}}$ and $\mathbf{B}^{\mathrm{feat}}$, are the same (either $\mathbf{T}_{t-1}^{\mathrm{feat}}$ or $\mathbf{D}_{t}^{\mathrm{feat}}$). For cross attention blocks, the input features are different (either $\mathbf{T}_{t-1}^{\mathrm{feat}}$ and $\mathbf{D}_{t}^{\mathrm{feat}}$ or $\mathbf{D}_{t}^{\mathrm{feat}}$ and $\mathbf{T}_{t-1}^{\mathrm{feat}}$, depending on the direction of cross-attention.) Self-attention blocks perform feature attenuation within a frame, therefore embedding spatial relationships through intra-frame interactions. Conversely, cross-attention blocks perform feature attenuation across frames, embedding temporal relationships through inter-frame interactions. Leveraging both self and cross attention provides a complete interaction modeling solution.
%------------------------------------------------------------------------
\\[0.2\baselineskip]
\noindent\textbf{Affinity Head}.
The interaction-aware track and detection features, $\mathbf{\tilde{T}}_{t-1}^{\mathrm{feat}}$ and $\mathbf{\tilde{D}}_t^{\mathrm{feat}}$, are used to generate an affinity matrix, $\mathbf{A}_t \in \mathbb{R}^{M \times N}$.  Every feature pair is concatenated to form affinity features, $\mathbf{F}_t \in \mathbb{R}^{M \times N \times 2C}$. The number of channels is reduced to 1 with a FFN, $\mathbf{G}_{\mathrm{aff}}$ (Channel Reduce in Figure~\ref{fig:architecture}), followed by a sigmoid layer to convert the predicted affinities into probabilities. The channel axis is flattened forming the affinity matrix, $\mathbf{A}_t$, where each element, $\mathbf{A}_t(m, n)$, represents the match confidence between a given track, $\mathbf{T}_{t-1}^m$, and detection, $\mathbf{D}_t^n$.
%------------------------------------------------------------------------
\subsection{3D Tracking} \label{sec:3dtracking}
%------------------------------------------------------------------------
The 3D tracking module is only used during inference, taking affinity estimates, $\mathbf{A}_t$, and 2D and 3D detections as input to produce tracks, $\mathbf{T}_t$. We adapt the EagerMOT~\cite{EagerMOT} framework due to the excellent performance of the two-stage data association. The framework performs track motion prediction, associates 3D and 2D detections with a fusion module, and separately associates 3D and 2D detections with tracks in two stages. Our adaptation modifies the first stage data-association, the track prediction and update, and introduces a track overlap rejection module.
%------------------------------------------------------------------------
\\[0.2\baselineskip]
\noindent\textbf{Fusion}.
We first match 3D and 2D detections directly following the fusion stage from the EagerMOT methodology~\cite{EagerMOT}. 3D detections are projected into the image, and greedily associated with 2D detections based on 2D IoU. Any matches below a 2D IoU threshold, $\tau_{fuse}$, are discarded.
%------------------------------------------------------------------------
\\[0.2\baselineskip]
\noindent\textbf{First Stage Data Association}.
We use our learned affinity, $\mathbf{A}_t$, for the first stage data association. In this stage all tracks, $\mathbf{T}_{t-1}$, are matched with all 3D object detections, $\mathbf{D}_t$, of the current frame. The Hungarian algorithm~\cite{Hungarian} is used find the optimal assignment matrix, $\mathcal{A}_t^*$, that minimizes the assignment affinity cost:

\begin{equation}
\begin{aligned}
    \mathcal{A}_t^* = &\underset{\mathcal{A}_t}{\argmin} \sum_{m,n}\mathbf{A}_t(m,n)\mathcal{A}_t(m,n)
    % \textrm{s.t.} &\quad  \sum_{m} \mathcal{A}_t(m,n)=1, \forall n\\
    % &\quad  \sum_{n} \mathcal{A}_t(m,n)=1, \forall m
\end{aligned}
\end{equation}
where each element $\mathcal{A}_t(m,n) \in \{0, 1\}$ and rows/columns are one-hot encoded. We maintain the match filtering strategy in EagerMOT~\cite{EagerMOT}, by removing assignments $(\mathbf{T}_{t-1}^m, \mathbf{D}_{t}^n)$ with scaled Euclidean distance~\cite{EagerMOT} greater than a threshold, $\tau_{3D}$. Experimentally, we find that the Hungarian algorithm outperforms greedy assignment with our learned affinity (see Table~\ref{tab:track-ablate}).
%------------------------------------------------------------------------
\\[0.2\baselineskip]
\noindent\textbf{Second Stage Data Association}.
The second stage data association directly follows the EagerMOT methodology~\cite{EagerMOT}. We use 2D detections without a matched 3D detection (from the fusion stage) and unmatched tracks (from the first stage) as input. 3D tracks are projected into the image plane, and are greedily associated with 2D detections based on 2D IoU. Matches below a 2D IoU threshold, $\tau_{2D}$, are removed.
%------------------------------------------------------------------------
\\[0.2\baselineskip]
\noindent\textbf{Track Prediction/Update}.
For track prediction, we discard the 3D Kalman filter and use the motion model used in CenterPoint~\cite{CenterPoint} as track prediction, which uses the latest velocity estimate from the detection as follows:

\begin{equation}
    \mathbf{p}_{t} = \mathbf{p}_{t-1} + d_t \mathbf{v}_{t-1},
\end{equation}
where $\mathbf{p}_t=(x_t, y_t)$ is the track centroid, $\mathbf{v}_{t-1}=(\dot{x}_{t-1}, \dot{y}_{t-1})$ is the predicted velocity of a detection and $d_t$ is the timestep duration at time $t$. Discarding the Kalman prediction equations removes the covariance predictions required for the Kalman update equations. Therefore, we simply assign the detection state as the latest measurement.
%------------------------------------------------------------------------
\\[0.2\baselineskip]
\noindent\textbf{Track Overlap Rejection}.
Due to the permissive structure of the two stage data association framework of EagerMOT, we note that duplicate tracks occur regularly in the output stream. Duplicate tracks often occur when ingesting detections~\cite{CenterPoint} with many duplicates with high recall, often with low confidence scores. Detections with low confidence could be filtered to remove most duplicates, but would result in reduced track coverage due to a loss of true positive detections. Rather, we detect duplicates by computing the 3D IoU between all tracks at each frame, and flag any pairs of the same class with a 3D IoU greater than a rejection threshold, $\tau_{\mathrm{rej}}$. For each pair, we reject the track with the lower age in order to reduce instances of identity switching.
%------------------------------------------------------------------------
\subsection{Training Loss} \label{sec:training-loss}
%------------------------------------------------------------------------
We treat the affinity matrix estimation as a binary classification problem where each element represents the probability of a match between any given track, $\mathbf{T}_{t-1}^m$, and detection, $\mathbf{D}_t^n$. We use the binary focal loss~\cite{FocalLoss} as supervision in order to deal with the positive/negative imbalance:
%------------------------------------------------------------------------
\begin{align}
   \label{eq:loss}
    L &= \frac{1}{M \cdot N}\sum_{m=1}^{M}\sum_{n=1}^{N}\mathrm{FL}(\mathbf{A}_t(m, n), \mathbf{\hat{A}}_t(m, n))
\end{align}
where $\mathbf{\hat{A}}_t$ is the affinity matrix label. To generate the affinity label, each track, $\mathbf{T}_{t-1}^m$, and detection, $\mathbf{D}_t^n$, is matched to ground truth boxes via 3D IoU to retrieve an identity label. Only matches greater than a threshold, $\tau_{\mathrm{gt}}$, are considered valid. We set $\mathbf{\hat{A}}_t(m, n) = 1$ for pairs $(\mathbf{T}_{t-1}^m, \mathbf{D}_t^n)$ with the same identity and $\mathbf{\hat{A}}_t(m, n) = 0$ for all remaining pairs.
%------------------------------------------------------------------------
\subsection{Data Augmentation} \label{sec:augmentation}
%------------------------------------------------------------------------
As the 3D object detector and \method~are trained on the same training splits, the quality of detections encountered at training time will be higher than at test time. Therefore, we apply two augmentations during training to reduce the gap between training and test detections.
%------------------------------------------------------------------------
\\[0.2\baselineskip]
\noindent\textbf{Positional Perturbation}.
To simulate positional error, we add a perturbation $(\delta_x, \delta_y)$ on the 2D position $(x, y)$ of detections by randomly sampling from a normal distribution:
\begin{align}
\label{eq:perturbation}
\begin{aligned}
    x &= x + \delta_x \\
    y &= y + \delta_y \\
\end{aligned}
\qquad
\begin{aligned}
   \delta_x \sim \mathcal{N}(0, \sigma_x^{2}) \\
   \delta_y \sim \mathcal{N}(0, \sigma_y^{2}) \\
\end{aligned}
\end{align}
%------------------------------------------------------------------------
\\[0.2\baselineskip]
\noindent\textbf{Detection Dropout}.
 We employ detection dropout to simulate missing detections. At each frame, a fraction $d$ of all detections are removed, where the fraction is randomly sampled from a uniform distribution $d \sim \mathcal{U}(d_{\mathrm{min}}, d_{\mathrm{max}})$.
%------------------------------------------------------------------------

%% file: content/results.tex
\section{Experimental Results} \label{experiments}
%-------------------------------------------------------------------------
\begin{table*}
\centering
\begin{tabular}{c|cc|ccccccc}
\toprule
\multirow{2}{*}{Method} & \multirow{2}{*}{AMOTA$\uparrow$} & \multirow{2}{*}{AMOTP$\downarrow$} & \multicolumn{7}{c}{AMOTA$\uparrow$} \\
 & & & Bicycle & Bus & Car & Motor & Ped & Trailer & Truck \\
 \midrule \midrule
AB3DMOT~\cite{AB3DMOT} & 15.1 & 1.501 & 0.0 & 40.8 & 27.8 & 8.1 & 14.1 & 13.6 & 1.3  \\
CenterPoint~\cite{CenterPoint} & 63.8 & 0.555 & 32.1 & 71.1 & 82.9 & 59.1 & 76.7 & 65.1 & 59.9   \\
ProbTrack~\cite{ProbMOTv2} & 65.5 & 0.617 & 46.9 & 71.3 &  \textbf{83.0} & 63.1 & 74.1 &  \textbf{65.7} & 54.6  \\
EagerMOT~\cite{EagerMOT} & 67.7 & \textbf{0.550} & 58.3 &  \textbf{74.1} & 81.0 & 62.5 & 74.4 & 63.6 & 59.7  \\
\midrule
\textbf{\method~(Ours)} & \textbf{68.7} & 0.560 &  \textbf{60.5} & 72.7 & 80.4 & \textbf{67.2} &  \textbf{77.0} & 62.9 &  \textbf{60.0} \\
\cellcolor{ImproveBlue}\textit{Improvement} & \cellcolor{ImproveBlue}\textit{+1.00} & \cellcolor{ImproveBlue}\textit{+0.01} & \cellcolor{ImproveBlue}\textit{+2.20} & \cellcolor{ImproveBlue}\textit{-1.40} & \cellcolor{ImproveBlue}\textit{-0.60} & \cellcolor{ImproveBlue}\textit{+4.70} & \cellcolor{ImproveBlue}\textit{+2.60} & \cellcolor{ImproveBlue}\textit{-0.70} & \cellcolor{ImproveBlue}\textit{+0.30} \\
\bottomrule
\end{tabular}
\caption{3D MOT results on the nuScenes test set~\cite{nuscenes2019}, reporting methods that use CenterPoint~\cite{CenterPoint} detections. We indicate the best result in \textbf{bold} and improvements relative to the next best method EagerMOT~\cite{EagerMOT}. Full results for \method~can be accessed \href{https://eval.ai/web/challenges/challenge-page/476/leaderboard/1321}{here}.}
\label{tab:nuscenes-test}
\end{table*}
%-------------------------------------------------------------------------
To demonstrate the effectiveness of \method~we present results on both the nuScenes 3D MOT benchmark~\cite{nuscenes2019} and the KITTI 3D MOT benchmark~\cite{Kitti}.

The nuScenes dataset~\cite{nuscenes2019} consists of dense urban environments, and is divided into 750 training scenes, 150 validation scenes, and 100 testing scenes. We compare \method~with existing methods on the test set, and use the validation set for ablations.

The KITTI dataset~\cite{Kitti} consists of mid-size city, rural and highway environments, and is divided into 21 training scenes and 29 testing scenes. We follow AB3DMOT~\cite{AB3DMOT} and divide the training scenes into a \textit{train} set (10 scenes) and a \textit{val} set (11 scenes), using the \textit{val} set for comparison.
%-------------------------------------------------------------------------
\\[0.2\baselineskip]
\noindent
\textbf{Input Parameters}.
The voxel grid is defined by a range and voxel size in 3D space. On nuScenes, we use $[-51.2, 51.2] \times [-51.2, 51.2] \times [-5, 3]$ (\si{\meter}) for the range and $[0.1, 0.1, 0.2]$ (\si{\meter}) for the voxel size. On KITTI,  we use $[0, 70.4] \times [-40, 40] \times [-3, 1]$ (\si{\meter}) for the range and $[0.05, 0.05, 0.1]$ (\si{\meter}) for the voxel size. We use leverage the same 3D and 2D detections as EagerMOT~\cite{EagerMOT}.
%-------------------------------------------------------------------------
\\[0.2\baselineskip]
\noindent
\textbf{Training and Inference Details}.
Our learned affinity estimation is implemented in PyTorch~\cite{Pytorch}. We use OpenPCDet~\cite{openpcdet2020} for the Voxel Backbone described in Section~\ref{sec:affinity-learning}. The network is trained on a NVIDIA Tesla V100 (32GB) GPU. The Adam optimizer~\cite{Adam} is used with an initial learning rate of 0.03 and is modified using the one-cycle learning rate policy~\cite{onecycle}. We train the model for 20 epochs on the nuScenes dataset~\cite{nuscenes2019} and 100 epochs on the KITTI dataset~\cite{Kitti} using a batch size of 4. We use previous detections $\mathbf{D}_{t-1}$ to represent tracks $\mathbf{T}_{t-1}$ during training. We use the 3D Kalman filter for track prediction on the KITTI~\cite{Kitti} dataset due to the lack of velocity estimates in KITTI detections.  The values $\alpha = 0.25$ and $\gamma = 2.0$ are used for the focal loss parameters in Equation~\ref{eq:loss}. We set $\tau_{\mathrm{rej}} = 0.6$ and $\tau_{\mathrm{gt}} = 0.55$ for the IoU thresholds in the track overlap rejection and affinity label generation steps, respectively. We set our detection perturbation standard deviations as $(\sigma_x, \sigma_y) = (0.01, 0.01)$, and set our detection drop fraction range as $(d_{\mathrm{min}}, d_{\mathrm{max}}) = (0.0, 0.2)$. We inherit all tracking related parameters from EagerMOT~\cite{EagerMOT}.
%-------------------.------------------------------------------------------
\subsection{nuScenes Dataset Results} \label{nuscenes-results}
%-------------------------------------------------------------------------
We use the official nuScenes 3D MOT evaluation~\cite{nuscenes2019} for comparison, reporting the AMOTA and AMOTP metrics. Table~\ref{tab:nuscenes-test} shows the results of \method~on the nuScenes test set~\cite{nuscenes2019}, compared to 3D MOT methods using CenterPoint~\cite{CenterPoint} detections for fair comparison. We observe that \method~outperforms previous methods by a large margin on overall AMOTA (+1.00\%) with a similar AMOTP (+0.01m).

\method~greatly outperforms previous methods on classes with smaller physical sizes, with considerable margins of +2.20\%, +4.70\%, and +2.60\% on the Bicycle, Motorcycle, and Pedestrian classes. We attribute the performance increase to the dense clustering of smaller objects that exists in the nuScenes dataset. Heuristic metrics often have trouble distinguishing between clustered objects (see Figure~\ref{fig:intro}), while our Interaction Transformer produces features with more discrimination leading to correct association. We reason heuristic metrics are sufficient on larger objects due to increased separation, as \method~lags behind prior methods on the Bus, Car, and Trailer classes with margins of -1.40\%, -0.60\%, and -0.70\%, and outperforms the Truck class (+0.30\%). We note \method~achieves 5 first place rankings.
%-------------------------------------------------------------------------
\subsection{KITTI Dataset Results} \label{kitti-results}
We follow AB3DMOT~\cite{AB3DMOT} for 3D MOT evaluation on the KITTI dataset~\cite{Kitti}. Table~\ref{tab:kitti} shows the results of \method~on the KITTI \textit{val} set~\cite{Kitti} for the Car class compared to 3D MOT methods. InterTrack achieves 1st place on the KITTI benchmark~\cite{AB3DMOT} with a 0.3\% margin on sAMOTA.
%-------------------------------------------------------------------------
\begin{table}
\small
\centering
\begin{tabular}{c|ccc}
\toprule
Method & sAMOTA$\uparrow$ & AMOTA$\uparrow$ & AMOTP $\uparrow$ \\
\midrule \midrule
 AB3DMOT$^\dagger$~\cite{AB3DMOT} & 91.8 & 44.3 & 77.4 \\
 GNN3DMOT$^\dagger$~\cite{AB3DMOT} & 93.7 & 45.3 & 78.1 \\
 EagerMOT~\cite{EagerMOT} & 94.9 & \textbf{48.8} & \textbf{80.4} \\
 \midrule
 \textbf{\method~(Ours)} & \textbf{95.2} & \textbf{48.8} & 80.3 \\
 \cellcolor{ImproveBlue}\textit{Improvement} & \cellcolor{ImproveBlue}\textit{+0.30} & \cellcolor{ImproveBlue}\textit{+0.00} & \cellcolor{ImproveBlue}\textit{-0.10} \\ 
\bottomrule
\end{tabular}
\caption{3D MOT results on the KITTI \textit{val} set~\cite{Kitti} for the Car class with the AB3DMOT~\cite{AB3DMOT} evaluation. $^\dagger$ indicates methods using 3D detections from PointRCNN~\cite{PointRCNN}.}
\label{tab:kitti}
\end{table}
%-------------------------------------------------------------------------
\subsection{Ablation Studies} \label{sec:ablations}
%------------------------------------------------------------------------
\begin{table}[t]
\small
\centering
\begin{tabular}{c|cccc}
\toprule
Exp & Model & AMOTA$\uparrow$ & AMOTP$\downarrow$ & AssA $\uparrow$  \\
\midrule \midrule
\newtag{1}{itm:fl1} & FFN & 63.4 & 0.749 & 48.2  \\
\newtag{2}{itm:fl2} & GNN & 66.3 & 0.688 & 51.1  \\
\newtag{3}{itm:fl3} & Trans & \textbf{72.1} & \textbf{0.566} & \textbf{56.2} \\
\bottomrule
\end{tabular}
\caption{Object Feature Learning Ablation on the nuScenes~\cite{nuscenes2019} validation set. We compare tracking performance when using a feed-forward network, a graph neural network, and a transformer for feature learning.}
\label{tab:feature-ablate}
\end{table}
We provide ablation studies on our network to validate our design choices. The results are shown in Tables~\ref{tab:feature-ablate} and~\ref{tab:track-ablate}.
%-------------------------------------------------------------------------
\begin{figure*}[t]
\begin{center}
\includegraphics[width=\textwidth]{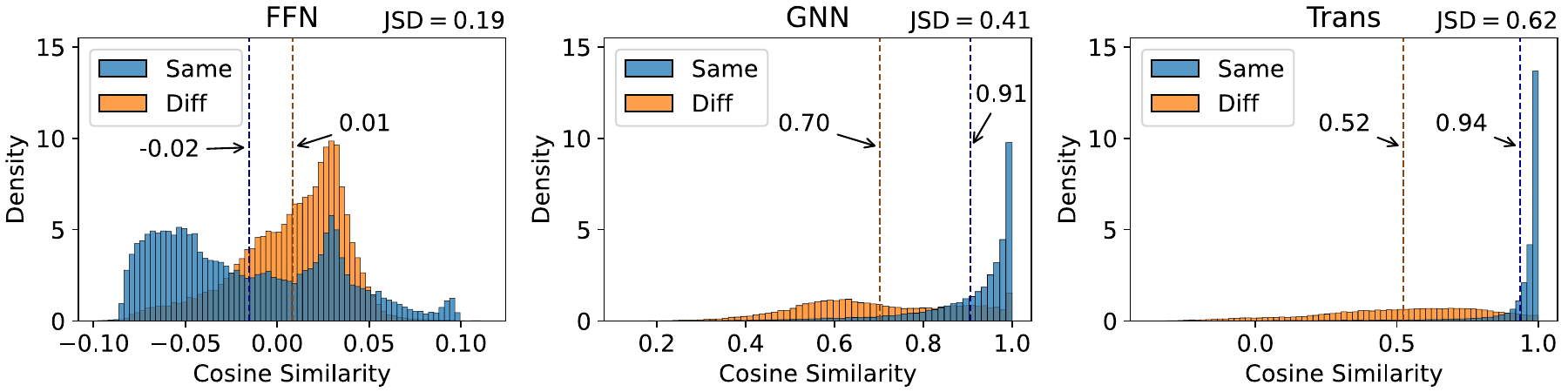}
\end{center}
\end{figure*}
%-------------------------------------------------------------------------
\begin{figure*}[t]
\begin{center}
\includegraphics[width=\textwidth]{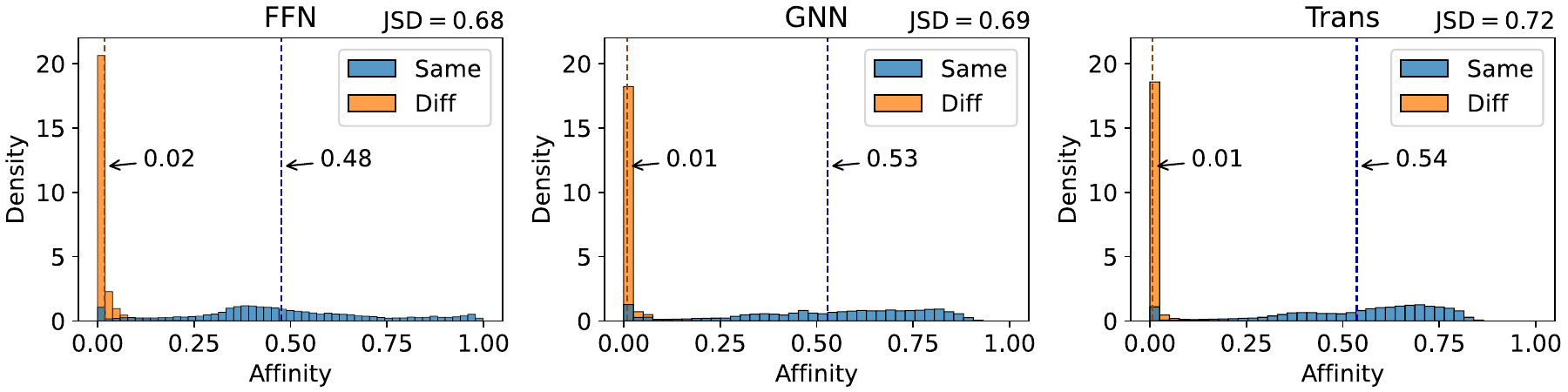}
\end{center}
   \caption{Discriminative Comparison for Feature Learning Models. We plot cosine similarity and affinity estimate $\mathbf{A}_t(m, n)$ distributions for both same and different object pairs for the FFN, GNN, and Transformer feature learning models. The distribution means (shown as dotted lines) and Jensen–Shannon divergence (JSD) between the distributions are shown.}
\label{fig:disc-compare}
\end{figure*}
%------------------------------------------------------------------------
\\[0.2\baselineskip]
\noindent
\textbf{Object Feature Learning}.
Table~\ref{tab:feature-ablate} shows the tracking performance when different feature learning models are used for affinity estimation. Specifically, we swap out the Interaction Transformer (Section~\ref{sec:affinity-learning}) with a feed-forward network (FFN) and a graph neural network (GNN). We evaluate the models using AMOTA, AMOTP, AssA on the nuScenes~\cite{nuscenes2019} validation set.

Experiment~\ref{itm:fl1} in Table~\ref{tab:feature-ablate} uses a FFN to learn object features independently, applying 8 Linear-BatchNorm-ReLU blocks. Experiment~\ref{itm:fl2} uses a GNN to add sparse interaction modeling. We follow the model architecture of GNN3DMOT~\cite{GNN3DMOT}, and only construct edges between objects that are (1) in different frames and (2) within 5 meters in 3D space. Adding the GNN leads to an improvement (+2.90\%, -0.061m, +2.90\%) when interactions are considered. We implement our own GNN for fair comparison instead of using the official GNN3DMOT~\cite{GNN3DMOT} results, which report poor performance as a result of using lower quality MEGVII~\cite{CBGS} detections. Experiment~\ref{itm:fl3} uses the Transformer model (described in Section~\ref{sec:affinity-learning}) for feature learning, with both self-attention and cross-attention layers. Leveraging the Transformer model leads to large performance improvements (+5.80\%, -0.122m, +5.10\%) due to the ability to model all object interactions, validating the use of the Transformer for 3D MOT.

We also directly compare the discrimination ability of the feature learning models in Figure~\ref{fig:disc-compare}. Specifically, we compute both feature similarity and affinities for all object pairs, and plot separate distributions for same and different object pairs. Same objects are object pairs with the same identity and different objects are intra-class object pairs with different identities. We assign ground truth object identities to objects using the 3D IoU matching strategy described in Section~\ref{sec:training-loss}. Feature similarity is evaluated using the cosine similarity between track and detection feature vectors, $\mathbf{\tilde{T}}_{t-1}^{\mathrm{feat}}, \mathbf{\tilde{D}}_t^{\mathrm{feat}}$, produced by the Interaction Transformer (see Section~\ref{sec:affinity-learning}). Affinity is directly taken from the affinity matrices, $\mathbf{A}_t$, produced by the Affinity Head (see Section~\ref{sec:affinity-learning}). 

We measure discrimination as the amount of separation between same and different object distributions, as increased separation reduces ambiguity when matching object pairs. As a measure of separation, we compare the distribution means and compute the Jensen–Shannon divergence (JSD).  The Jensen–Shannon divergence (JSD) is used over the standard Kullback–Leibler (KL) divergence due to the symmetrical and always finite output of the metric.

As in Table~\ref{tab:feature-ablate}, we observe improvements with increased interaction modeling in Figure~\ref{fig:disc-compare}. Sparse interaction modeling with the GNN shows an improvement over independent feature extraction with the FFN, with increased mean difference (+0.24) and JSD difference (+0.22) for feature similarities and an increased mean difference (+0.06) and JSD difference (+0.01) for affinities. Adding complete interaction modeling with the Transformer leads to an increased mean difference (+0.21) and JSD difference (+0.21) for feature similarities and an increased mean difference (+0.01) and JSD difference (+0.03) for affinities, validating the use of the Transformer for discriminative feature and affinity estimation. Based on Table~\ref{tab:feature-ablate} and Figure~\ref{fig:disc-compare}, we note that discrimination increases lead to increases in AMOTA, indicating the importance of object discrimination for tracking performance.

Further, we observe different characteristics between the cosine similarity and affinity distributions in Figure~\ref{fig:disc-compare}, which we attribute to the difficulty and frequency imbalance between positive (same object) and negative (different object) matches in the nuScenes dataset~\cite{nuscenes2019}. For the GNN and Transformer, we note a dense clustering of same object feature similarities near 1 (maximum similarity) and a larger spread of different object similarities. We reason a high difficulty of positive matches would require a high feature similarity to identify correctly, while a negative match could be identified with a range of lower similarities. Conversely, we observe a dense clustering of different object affinities near 0 (minimum affinity), attributed to the large amount of easy, negative matches to identify in the dataset. We note a larger spread of same object affinities which we attribute to the challenging nature of positive match identification.
%-------------------------------------------------------------------------
\begin{table}[t]
\small
\centering
\begin{tabular}{c|c|ccccc}
\toprule
Exp & MA & Rej & $\mathbf{A}_t$ & Pred & AMOTA$\uparrow$ & AMOTP$\downarrow$ \\
\midrule \midrule
\newtag{1a}{itm:ab1a} & \multirow{4}{*}{G} & & Heuri & KF & 71.2 & 0.569 \\
\newtag{2a}{itm:ab2a} & & \checkmark & Heuri & KF & 71.3 & 0.568 \\
\newtag{3a}{itm:ab3a} & & \checkmark & Learn & KF & 71.4 & 0.586\\
\newtag{4a}{itm:ab4a} & & \checkmark & Learn & Vel & 71.6 & 0.585\\
\midrule
\newtag{1b}{itm:ab1b} & \multirow{4}{*}{H} & & Heuri & KF & 63.8 & 0.635\\
\newtag{2b}{itm:ab2b} & & \checkmark & Heuri & KF & 65.8 & 0.628 \\
\newtag{3b}{itm:ab3b} & & \checkmark & Learn & KF & 71.8 & 0.573 \\
\newtag{4b}{itm:ab4b} & & \checkmark & Learn & Vel & \textbf{72.1} & \textbf{0.566} \\
\bottomrule
\end{tabular}
\caption{3D Tracking Ablation on the nuScenes~\cite{nuscenes2019} validation set. MA indicates the matching algorithm in the \nth{1} stage data association, where G is Greedy and H is Hungarian. Rej indicates track overlap rejection. $\mathbf{A}_t$ indicates the \nth{1} stage affinity estimation method, where Heuri indicates the EagerMOT~\cite{EagerMOT} heuristic and Learn indicates our learned approach.  Pred indicates the prediction model, where KF indicates the Kalman filter and Vel indicates the CenterPoint~\cite{CenterPoint} velocity-based motion model.}
\label{tab:track-ablate}
\end{table}
%------------------------------------------------------------------------
\\[0.2\baselineskip]
\noindent
\textbf{Track Rejection}.
Experiment~\ref{itm:ab1a} in Table~\ref{tab:track-ablate} shows the tracking performance of the EagerMOT~\cite{EagerMOT} baseline without any additions, while Experiment~\ref{itm:ab1b} shows the performance when using the Hungarian algorithm~\cite{Hungarian} in the first stage data association. We add a track rejection based on overlap (see Section~\ref{sec:3dtracking}) in Experiments~\ref{itm:ab2a} and~\ref{itm:ab2b}, leading to improvements on the Greedy (+0.10\%, -0.001m) and Hungarian (+2.00\%, -0.007m) trackers on AMOTA and AMOTP. We attribute the performance increase to the removal of duplicated trajectories estimated by our tracker.
%------------------------------------------------------------------------
\\[0.2\baselineskip]
\noindent
\textbf{Affinity Matrix Estimation}.
Experiments~\ref{itm:ab3a} and~\ref{itm:ab3b} in Table~\ref{tab:track-ablate} replace the heuristic affinity matrix estimation method used in EagerMOT~\cite{EagerMOT} with the learned affinity $\mathbf{A}_t$ estimation method outlined in Section~\ref{sec:affinity-learning}. We note improvements (+0.10\%, -0.002m and +6.00\%, -0.055m), which we attribute the increased discrimination of our feature interaction mechanism.
%------------------------------------------------------------------------
\\[0.2\baselineskip]
\noindent
\textbf{Track Prediction/Update}.
Experiments~\ref{itm:ab4a} and~\ref{itm:ab4b} in Table~\ref{tab:track-ablate} replace the Kalman filter with the estimated velocity prediction and assignment update strategy outlined in Section~\ref{sec:3dtracking}). We observe a performance improvement (+0.20\%, -0.001m and +0.30\%, -0.007m) over the Kalman filtering approach, which we credit to the low accuracy of velocity propagation when filtering, especially when operating at the low 2 Hz update rate of the nuScenes dataset. When ingesting detections without velocity estimates or when operating at a higher update rate, \method~may benefit from a Kalman filtering approach.
%------------------------------------------------------------------------
\\[0.2\baselineskip]
\noindent
\textbf{Data Association Method}.
Table~\ref{tab:track-ablate} shows the performance for all experiments when both the Greedy and Hungarian algorithms are used in the first stage data association. We observe that the Greedy algorithm (Experiments~\ref{itm:ab1a} and~\ref{itm:ab2a}) offers higher performance (+7.40\%, -0.066m and +5.50\%, -0.060m) over the Hungarian algorithm (Experiments~\ref{itm:ab1b} and~\ref{itm:ab2b}) when a heuristic affinity metric is used. We attribute this performance difference to the globally optimal solution that the Hungarian algorithm provides~\cite{Hungarian} which considers all affinity scores for all matches. We reason that a heuristic affinity metric produces more incorrect scores. For a majority of matches, the greedy approach only considers a subset of affinities in which incorrect affinity scores are ignored.

We observe better performance (+0.40\%, -0.013m and +0.50\%, -0.019m) for the Hungarian algorithm (Experiments~\ref{itm:ab3b} and~\ref{itm:ab4b}) compared to the Greedy algorithm (Experiments~\ref{itm:ab3a} and~\ref{itm:ab4a}) when the affinity is learned. Similarly, we attribute this performance difference to the globally optimal solution from the Hungarian algorithm that takes advantage of our improved affinity estimates. The combination of learned affinity estimation and Hungarian algorithm leads to the best results, validating its use for our method.

Additionally, we observe larger performance gains (+2.00\%, -0.007m and +6.00\%, -0.055m) for the Hungarian methods from Experiments~\ref{itm:ab1b} to~\ref{itm:ab3b} compared to gains (+0.10\%, -0.001m and +0.10\%, -0.002m) for the Greedy methods from Experiments~\ref{itm:ab1a} to~\ref{itm:ab3a}. We credit the performance gain difference to the high baseline performance (71.2\%, 0.569m) when using the Greedy algorithm (Experiment~\ref{itm:ab1a}),  in which further additions have less effect.
%------------------------------------------------------------------------

%% file: content/conclusion.tex
\section{Conclusion} \label{Conclusions}
We have presented \method, a novel 3D multi-object tracking method that extracts discriminative track and detection features for data association via end-to-end learning. The Transformer is used as a feature interaction mechanism, shown to be effective for increasing 3D MOT performance due to its integration of complete interaction modeling. We outline updates to the EagerMOT~\cite{EagerMOT} tracking pipeline leading to improved performance when paired with the Interaction Transformer. Our contributions lead to a \nth{1} place ranking on the nuScenes 3D MOT benchmark~\cite{nuscenes2019} among methods using CenterPoint~\cite{CenterPoint} detections.

%% file: content/appendix.tex
\appendix
\gdef\thesection{Appendix \Alph{section}}
%------------------------------------------------------------------------
\section{Additional Results}
We note that \method~reports more ID switches than comparable methods. Table~\ref{tab:secondary-metrics} reports the ID switch results on the nuScenes validation set for the EagerMOT baseline and \method. At first glance, the results would suggest that \method~poorly performs data association, however we argue that the nuScenes ID switch metric is unreliable for comparison between methods. The ID switch metric is computed at a single confidence threshold, which is selected as the threshold that results in the highest MOTA~\cite{nuscenes2019}. The confidence threshold operating point can differ between methods, resulting in an unfair comparison. For example, if the confidence threshold for one method is higher, fewer track predictions will be considered for evaluation, which will result in fewer ID switches from simply evaluating at a different confidence threshold. When analysing the association performance between trackers we recommend using the more recent HOTA~\cite{HOTA} metric. HOTA is advantageous since it is evaluated over a range of confidence thresholds and can decompose the evaluation of the tracking task into Detection Accuracy (DetA) and Association Accuracy (AssA). Under the HOTA evaluation scheme \method~clearly demonstrates that it outperforms the EagerMOT baseline by improving AssA by $1.7\%$.

Additionally, we ablate our use of a feed-forward network in the Affinity Head with traditional metrics for affinity computation in Table~\ref{tab:aff}. We compare against Cosine Similarity and Inner Product metrics computed between detection and track feature vectors at the input to the Affinity Head. We note strong performance of the FFN relative to traditional metrics, with improvements of +4.00\%, -0.072, and -1014 on AMOTA, AMOTP, and IDS on the nuScenes validation set.
%------------------------------------------------------------------------
\section{Potential Negative Impact}
We identify surveillance as the most significant negative application of InterTrack. Our work is designed for autonomous vehicle applications, however this technology could potentially be applied for 3D tracking in other applications such as traffic surveillance, where the various objects such as vehicles, pedestrians, cyclists are tracked. Our work only tracks the 3D pose information, but one could imagine more information being tracked including facial identities, license plates, and specific locations that are visited frequently. We note that all 3D MOT methods have the potential for use in surveillance.
%------------------------------------------------------------------------
\section{Limitations}
%------------------------------------------------------------------------
\noindent\textbf{Sensor Dependency \& Distribution Shift}.
Our learned Interaction Transformer introduces a dependency on the raw LiDAR data stream that does not exist for trackers who operate solely on detector output. While this dependency enables InterTrack to extract more powerful feature representations for objects, it limits the flexibility of our model. The Interaction Transformer is sensitive to the detection and LiDAR point cloud input distributions. We expect that our model would not generalize to different LiDAR sensors without re-training. Performance may also degrade if the tracker observes scenes with properties significantly different from those observed during training. Neither of these limitations have been experimentally verified, however, future work should explore generalization of tracking methods beyond single sensor sets or datasets.
%------------------------------------------------------------------------
\\[0.2\baselineskip]
\noindent\textbf{Interactions}.
Our work specifically targets tracking applications for densely occupied urban environments, in which interaction information can be helpful due to close proximity. For sparsely distributed  objects, we reason that heuristic affinities are sufficient and including interaction information is less beneficial (see Section 4.1). We expect interaction modeling to be useful for 3D MOT in urban environments and less so for suburban and rural applications.
%------------------------------------------------------------------------
\begin{table}[t]
\small
\centering
\begin{tabular}{c|cccc}
\toprule
Model & HOTA $\uparrow$ & DetA $\uparrow$ & AssA $\uparrow$ & IDS $\downarrow$  \\
\midrule \midrule
EagerMOT~\cite{EagerMOT} & 24.9 & \textbf{13.5} & 54.5 &  \textbf{899}\\
\textbf{InterTrack (Ours)} & \textbf{25.4} & \textbf{13.5} & \textbf{56.2} & 1019 \\
\bottomrule
\end{tabular}
\caption{Comparison of Data Association Metrics on the nuScenes \cite{nuscenes2019} validation set. We compare \method~with the EagerMOT~\cite{EagerMOT} baseline.}
\label{tab:secondary-metrics}
\end{table}

\begin{table}
\small
\centering
\begin{tabular}{c|ccc}
\toprule
Affinity Estimation & AMOTA$\uparrow$ & AMOTP $\downarrow$ & IDS $\downarrow$\\
\midrule \midrule
Inner Product & 66.2 & 0.689 & 3561\\
Cosine Similarity & 68.1 & 0.638 & 2033 \\
FFN (Ours) & \textbf{72.1} & \textbf{0.566} & \textbf{1019} \\
\bottomrule
\end{tabular}
\caption{Affinity Estimation Ablation on the nuScenes val set.}
\label{tab:aff}
\end{table}